\documentclass[11pt,letterpaper]{article}
\usepackage{naaclhlt2015}
\usepackage{times}
\usepackage{latexsym}
\usepackage{graphicx}

\usepackage[font=small]{caption}
\usepackage[font=footnotesize]{subcaption}
\usepackage{times}
\usepackage{latexsym}
\usepackage{amsmath}
\usepackage{multirow}
\usepackage{url}

\usepackage{amsmath,amssymb,amsbsy}
\usepackage{multirow}
\usepackage{color}
\usepackage{graphicx}

\usepackage{pgfplots}
\usepackage{algorithmicx}
\usepackage{algorithm}
\usepackage{algpascal}
\usepackage{algc}
\usepackage{algcompatible}
\usepackage{algpseudocode}

\newcommand{\cN}{\mathcal{N}}

\newcommand{\bw}{\mathbf{w}}

\newcommand{\bU}{\mathbf{U}}
\newcommand{\bV}{\mathbf{V}}

\title{Inferring Missing Entity Type Instances for Knowledge Base Completion: \\New Dataset and Methods}

\author{Arvind Neelakantan%
\thanks{\hspace{2 mm}Most of the research conducted during summer internship at Microsoft.}\\
Department of Computer Science \\
University of Massachusetts, Amherst \\ Amherst, MA, 01003 \\
	    {\tt arvind@cs.umass.edu}
	  \And
	Ming-Wei Chang\\
Microsoft Research\\
1 Microsoft Way\\
Redmond, WA 98052, USA\\
  {\tt  minchang@microsoft.com}}

\date{}

\begin{document}
\maketitle
%\footnote[*]{Most of the research conducted during summer internship at Microsoft.}
\begin{abstract}
Most of previous work in knowledge base (KB) completion has focused
on the problem of relation extraction. In this work, we focus on the task of  inferring missing entity
type instances in a KB, a fundamental task for KB competition yet
receives little attention.

Due to the novelty of this task, we construct a large-scale dataset and design an 
automatic evaluation methodology.  Our knowledge base completion method uses information within the existing KB and  external information from Wikipedia. We show that individual methods  trained with  a \emph{global} objective that considers unobserved cells from both
the entity and the type side gives consistently higher quality predictions compared to baseline methods. We also perform manual evaluation on
a small subset of the data to verify the effectiveness of our
knowledge base completion methods and the correctness of our proposed automatic evaluation method.
\end{abstract}

\section{Introduction}
\label{sec:introduction}
%\mnote{we also need to convince them global measurement is very important here., adding some numbers to convince them
% is very imbalance so we need to have global}
% \mnote{replace the picture the one that has the missing information}
\begin{figure*}
\centering
\includegraphics[scale=0.57]{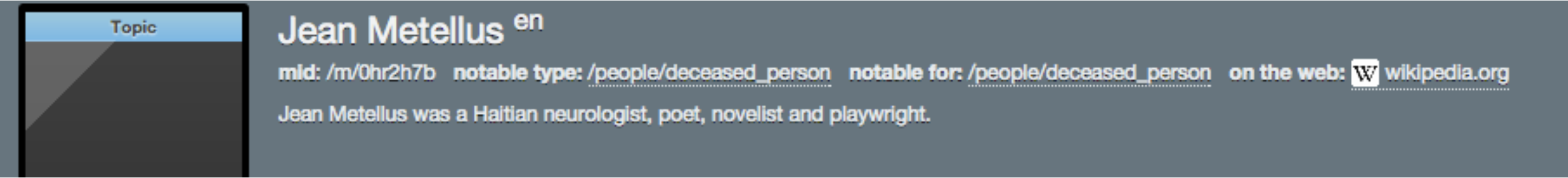}
\caption{\small Freebase description of \emph{Jean Metellus} can be used to infer that the entity has the type \emph{/book/author}. This missing fact is found by our algorithm and is still missing in the latest version of Freebase when the paper is written.}
\label{fd}
\end{figure*}
There is now increasing interest in the construction of knowledge
bases like \emph{Freebase} \cite{freebase} and \emph{NELL} \cite{NELL}
in the natural language processing community. KBs contain facts such
as \emph{Tiger Woods} is an \emph{athlete}, and \emph{Barack Obama} is
the \emph{president of} \emph{USA}. However, one of the main drawbacks
in existing KBs is that they are incomplete and are missing important
facts \cite{WGMSGL14}, jeopardizing their usefulness in downstream
tasks such as question answering. This has led to the task of
completing the knowledge base entries, or Knowledge Base Completion
(KBC) extremely important.

 In this paper, we address an important subproblem of knowledge base
 completion--- inferring missing entity type instances. Most of
 previous work in KB completion has only focused on the problem of
 relation extraction \cite{distant_supervision,rescal,transe,limin}.
 Entity type information is crucial in  KBs and  is  widely used in many NLP  tasks such as relation extraction \cite{trescal},
 coreference resolution \cite{ratinov,coref}, entity linking
 \cite{entity_linking}, semantic parsing \cite{ccg,dcs} and question
 answering \cite{subgraph,ie}. For example, adding entity type information improves relation extraction by 3\% \cite{trescal} and entity linking by 4.2 F1 points \cite{elnew}. Despite their importance, there is surprisingly little  previous work on this problem and, there are no 
 datasets publicly available for evaluation.

We construct a large-scale dataset for the task of inferring missing entity type instances in a KB. Most of previous KBC datasets \cite{distant_supervision,limin} are constructed using a single snapshot of the KB and methods are evaluated on a subset of facts that are hidden during training. Hence, the methods could be potentially evaluated by their ability to predict \emph{easy} facts that the KB already contains. Moreover, the  methods are not directly evaluated on their ability to predict missing facts. To overcome these drawbacks we construct the train and test data using two snapshots of the KB and evaluate the methods on predicting facts that are added to the more recent snapshot, enabling a more realistic and challenging evaluation. 

Standard evaluation metrics for KBC methods are generally {\em type-based} \cite{distant_supervision,limin}, measuring the quality of the predictions  by aggregating scores computed within a type. This  is not ideal because: (1) it treats every entity type equally not    considering the distribution of types, (2) it does not measure the ability of the methods to rank predictions across types. Therefore, we additionally use a
global evaluation metric, where the quality of predictions is measured within
and across types, and also accounts for the high variance in type distribution. In our experiments, we show that models trained with negative examples from the entity side perform better on type-based metrics, while when trained with negative examples from the type side perform better on the global metric.

In order to design methods that can rank predictions {\em both} within
and across entity (or relation) types, we propose a {\em global
  objective} to train the models. Our proposed method combines the
advantages of previous approaches by using negative examples from both
the entity and the type side. 
When considering the same number of negative examples, we find that the linear classifiers and the  low-dimensional  embedding models trained with the global objective produce better quality ranking within and across entity types when compared to training with negatives examples only from entity or type side.    Additionally compared to prior methods, the model
trained on the proposed global objective can more reliably suggest
confident entity-type pair candidates that could be added into the
given knowledge base.
% Prior to
% this work, there are {\em type-based approaches}, which sample negative examples only from the type side
% ~\cite{distant_supervision,limin} and {\em entity-based approaches}, that sample negative examples only from the entity side~\cite{fine-grained}. The global training objective approach combines the advantages of these 
% two approaches by using negative examples from both the entity and the type side.

Our contributions are summarized as follows:
\begin{itemize}
\item We develop an evaluation framework comprising of methods for dataset construction and evaluation metrics to
  evaluate KBC approaches for missing entity type instances. The dataset and evaluation scripts are publicly available at {\footnotesize \url{http://research.microsoft.com/en-US/downloads/df481862-65cc-4b05-886c-acc181ad07bb/default.aspx}}.
\item We propose a global training objective for KBC methods. The experimental results show that both linear classifiers and low-dimensional embedding models achieve best overall performance when trained with the global objective function.

\item We conduct extensive studies on models for inferring missing
  type instances studying the impact of using various features and
  models.
\end{itemize}

\section{Inferring Entity Types}
\label{sec:information-sources}
We consider a KB $\Lambda$ containing entity type information of the form $(e,
t)$, where $e \in E$ ($E$ is the set of all entities) is an entity in the KB with type $t \in T$ ($T$ is the set of all types).  For
example, $e$ could be \emph{Tiger Woods} and $t$ could be
\emph{sports athlete}. As a single entity can have multiple types, entities in Freebase
often miss some of their types. The aim of this work is to infer missing entity type instances in the KB. 
Given an unobserved fact (an entity-type pair) in the training data $(e, t) \not\in \Lambda$ where  entity $e \in E$ and type $t \in T$, the task is to infer whether the KB currently misses the  fact, i.e., infer whether $(e, t) \in \Lambda$. We consider entities in the intersection of Freebase and Wikipedia in our experiments.

\subsection{Information Resources}
Now, we describe the information sources used to construct the feature representation of an entity to infer its types. We use information in Freebase and external information from Wikipedia to complete the KB.
\begin{itemize}
\item{{\bf Entity Type Features}: The entity types observed in the training data can be a useful signal to infer missing entity type instances. For example, in our snapshot of \emph{Freebase}, it is not uncommon to find an entity with the type \emph{/people/deceased\_person} but missing the type \emph{/people/person}.  
}
\item{{\bf Freebase Description}: Almost all entities in \emph{Freebase} have a short one paragraph description of the entity. Figure \ref{fd}  shows the \emph{Freebase} description of \emph{Jean Metellus} that can be used to infer the type \emph{/book/author} which \emph{Freebase} does not contain as the date of writing this article.
\begin{figure*}
\centering
\includegraphics[scale=0.58]{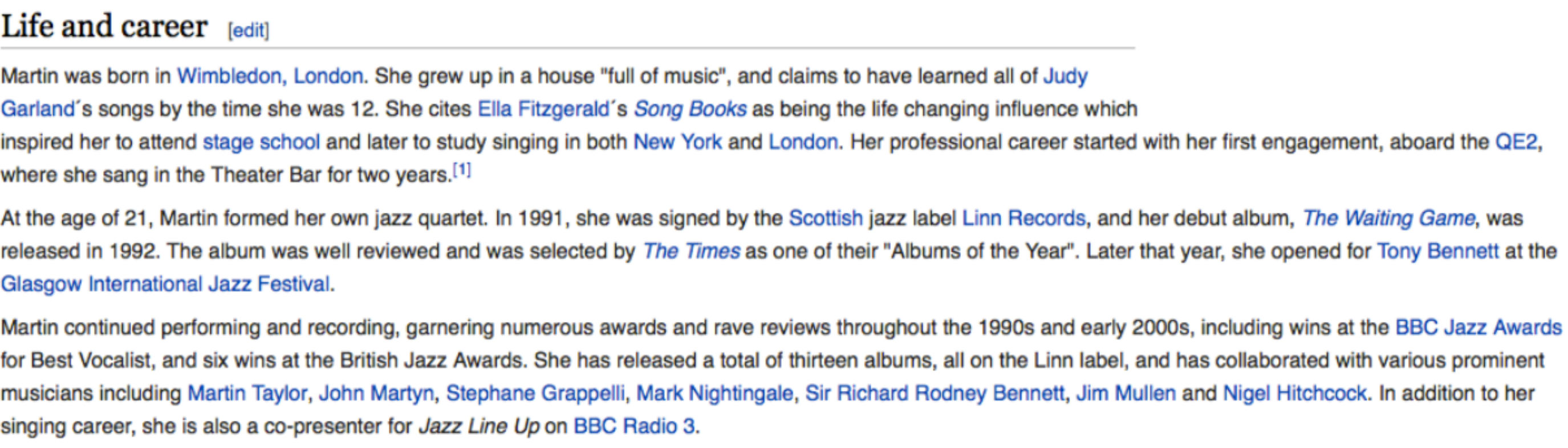}
\caption{\small A section of the \emph{Wikipedia} article of \emph{Claire Martin} which gives clues that entity has the type \emph{/award/award\_winner}. This currently missing fact is also found  by our algorithm. }
\label{wiki}
\end{figure*}
}

\item{{\bf Wikipedia}: As external information, we include the \emph{Wikipedia} full text article of an entity in its feature representation. We consider entities in Freebase that have a link to their Wikipedia article.  The  \emph{Wikipedia} full text of an entity gives several clues to predict it's entity types.  For example, Figure \ref{wiki} shows a section of the \emph{Wikipedia} article of \emph{Claire Martin} which gives clues to infer the type \emph{/award/award\_winner} that \emph{Freebase} misses.}
\end{itemize}

\section{Evaluation Framework}
\label{sec:eval-meth}
In this section, we propose an evaluation methodology for the
task of inferring missing entity type instances in a KB. While we
focus on recovering entity types, the proposed framework
can be easily adapted to relation extraction as well.

First, we discuss our two-snapshot dataset construction strategy. Then we motivate the importance of evaluating KBC algorithms globally and describe the evaluation metrics we employ.
\subsection{Two Snapshots Construction}

In most previous work on KB completion to predict missing relation facts
\cite{distant_supervision,limin}, the methods are evaluated on a
subset of facts from a {\em single} KB snapshot, that are hidden while
training.  However, given that the missing entries are usually
selected randomly, the distribution of the selected unknown entries could be
very different from the actual missing facts distribution. Also, since any fact could be potentially used for evaluation, the methods could be evaluated on their ability to predict easy facts that are already present in the KB.

To overcome this drawback, we construct our train and test  set by
considering {\em two} snapshots of the knowledge base. The {\em
  train} snapshot is taken from an earlier time without special
treatment. The {\em test} snapshot is taken from a later period, and a
KBC algorithm is evaluated by its ability of recovering newly added
knowledge in the test snapshot. This enables the methods to be
directly evaluated on facts that are missing in a KB snapshot. Note
that the facts that are added to the test snapshot, in general, are
more subtle than the facts that they already contain and predicting
the newly added facts could be harder. Hence, our approach enables a
more realistic and challenging evaluation setting than previous work.

We use manually constructed  \emph{Freebase}  as the KB in our experiments.  Notably, \newcite{trescal} use a two-snapshot strategy for constructing a dataset for relation extraction  using automatically constructed  \emph{NELL} as their KB. The new facts that are added to a KB by an automatic method may not have all the characteristics that make the two snapshot strategy more advantageous.

We construct our train snapshot $\Lambda_0$ by taking the \emph{Freebase}
snapshot on $3^{rd}$ September, 2013 and consider entities that have a
link to their Wikipedia page.
KBC algorithms are evaluated by their ability to predict facts that were
added to the $1^{st}$ June, 2014 snapshot of \emph{Freebase} $\Lambda$. To get
negative data, we make a closed world assumption treating any
unobserved instance in \emph{Freebase} as a negative
example. Unobserved instances in the \emph{Freebase}
snapshot on $3^{rd}$ September, 2013 and $1^{st}$ June, 2014 are used
as negative examples in training and testing respectively.\footnote{Note that some of the negative instances used in training could be positive instances in test but we do not remove them during training.}

The positive instances in the test data ($\Lambda - \Lambda_0$) are facts that are newly added to the test snapshot $\Lambda$.
Using the entire set of negative examples in the test data is impractical due to the large number of  negative examples. To avoid this we only add the negative types of entities that have at least one new fact in the test data. Additionally, we add a portion of the negative examples for entities which do not have new fact in the test data and that were unused during training. This makes our dataset quite challenging since the number of negative instances is much larger than the number of positive instances in the test data.

It is  important to note that the goal of this work is not to
predict facts that emerged between the time period of the train and test
snapshot\footnote{In this work, we also do not aim to correct existing
  false positive errors in \emph{Freebase}}. For example, we do not
aim to predict the type \emph{/award/award\_winner} for an entity that
won an award after $3^{rd}$ September, 2013. Hence, we use
the \emph{Freebase} description in the training data snapshot and
\emph{Wikipedia} snapshot on $3^{rd}$ September, 2013 to get the
features for entities.

One might worry that the new snapshot might contain a significant amount of
emerging facts so it could not be an effective way to evaluate the KBC
algorithms. Therefore, we examine the difference between the training
snapshot and test snapshot manually and found that this is likely not the case.
For example, we randomly selected $25$ \emph{/award/award\_winner}
instances that were added to the test snapshot and found that all of
them had won at least one award before $3^{rd}$ September, 2013.

Note that while this automatic evaluation is closer to the real-world
scenario, it is still not perfect as the new KB snapshot is still incomplete. Therefore, we also perform human evaluation
on a small dataset to verify the effectiveness of our approach.

\subsection{Global Evaluation Metric}
\emph{Mean average precision} (MAP) \cite{map} is now commonly used to
evaluate KB completion methods \cite{distant_supervision,limin}. MAP
is defined as the mean of \emph{average precision} over all entity
(or relation) types. MAP treats each entity type equally (not explicitly accounting for their distribution).
However, some types occur much more frequently than others. For example, in our large-scale experiment with $500$ entity types, there are many entity types with only $5$ instances in the test set while the most frequent  entity type has tens of thousands of missing instances. Moreover,  MAP only measures the ability of the methods to correctly rank predictions within a type.

To account for the high variance in the distribution of entity types and measure the ability of the methods to correctly rank predictions  across types  we use  global average precision (GAP) (similarly to
micro-F1) as an additional evaluation metric for KB completion. We convert the
multi-label classification problem to a binary classification problem
where the label of an entity and type pair is true if the entity has
that type in \emph{Freebase} and false otherwise. GAP is the average
precision of this transformed problem which can measure the ability of
the methods to rank predictions both within and across entity
types.

 Prior to us, \newcite{transe}  use mean reciprocal rank as a global evaluation metric for a KBC task. We use average precision instead of  mean reciprocal rank since MRR could be biased to the top predictions of the method \cite{WGMSGL14}

While GAP captures global ordering, it would be beneficial to measure the quality of the top $k$ predictions of the model for bootstrapping and active learning scenarios \cite{active_learning,boot}. We report G@k, GAP measured on the top $k$
predictions (similarly to \emph{Precision@k} and \emph{Hits@k}). This metric can be reliably used to measure the overall
quality of the top $k$ predictions.

%%% Local Variables:
%%% mode: latex
%%% TeX-master: "paper"
%%% TeX-PDF-mode: t
%%% End:

\section{Global Objective for Knowledge Base Completion}
\label{sec:approach}
We describe our approach for predicting missing entity types in a KB
in this section. While we focus on recovering entity types in this
paper, the methods we develop can be easily extended to other KB
completion tasks.

\subsection{Global Objective Framework}
During training, only positive examples are observed in KB completion
tasks. Similar to previous work
\cite{distant_supervision,transe,limin}, we get negative training
examples by treating the unobserved data in the KB as negative
examples.  Because the number of unobserved examples is
much larger than the number of facts in the KB, we follow previous
methods and sample few unobserved negative examples for every positive
example.
%  The class imbalance problem can impact
% the performance of the classifiers, and also results in
% long training time.

Previous methods largely neglect the sampling methods on
unobserved negative examples. The proposed global object framework allows us to systematically
study the effect of the different sampling methods to get negative data, as the performance of the model for different evaluation metrics does depend on the sampling method.
%As sExperimental results Section~\ref{sec:experiments} indicate the performance of the model for different evaluation metrics does depend on the choice of the sampling method.
%  We also show that the existing methods are  special cases of the
%global objective framework and it combines the benefits of the two existing orthogonal approaches.

We consider a training snapshot of the KB $\Lambda_0$, containing facts of the form $(e, t)$
where $e$ is an entity in the KB with type $t$. Given a fact $(e, t)$ in the KB, we
consider two types of negative examples constructed from the following two sets: $\cN_E (e,t)$ is the
``negative entity set'', and $\cN_T (e,t)$ is the ``negative
type set''.  More precisely,

\begin{equation*}
 \cN_E (e,t) \subset \{e'| e' \in E, e' \neq e, (e',t) \notin \Lambda_0\},
\end{equation*}
and

\begin{equation*}
 \cN_T (e,t) \subset \{t'| t'\in T,  t' \neq t, (e,t') \notin \Lambda_0\}.
\end{equation*}

Let $\theta$ be the model parameters, $m = |\cN_E (e,t)|$ and $n = |\cN_T (e,t)|$ be the number of negative examples and types considered for training respectively.  For each entity-type pair
$(e,t)$, we define the scoring function of our model as $s(e,t |
\theta)$.\footnote{We often use $s(e,t)$ as an abbreviation of
  $s(e,t|\theta)$ in order to save space.}  We define two loss functions one using negative entities and the other using negative types:
\begin{equation*}
 L_E (\Lambda_0, \theta) = \!\!\!\!\sum_{(e,t)\in\Lambda_0, e' \in \cN_E(e,t)} \!\!\!\! [s(e',t) - s(e,t) +1]_+^k,
\end{equation*}
and
\begin{equation*}
 L_T (\Lambda_0, \theta) = \!\!\!\!\sum_{(e,t)\in\Lambda_0, t' \in \cN_T(e,t)} \!\!\!\! [s(e,t') - s(e,t) +1]_+^k,
\end{equation*}
where $k$ is the power of the loss function ($k$ can be 1 or 2), and the
function $[\cdot]_+$ is the hinge function.

The global objective function is defined as
\begin{equation}
  \label{eq:general_obj}
\min_{\theta} Reg(\theta) + CL_T(\Lambda_0, \theta) + CL_E(\Lambda_0, \theta),
\end{equation}
where $Reg(\theta)$ is the regularization term of the model,
and $C$ is the regularization parameter.  Intuitively, the parameters
$\theta$ are estimated to rank the observed facts above the negative
examples with a margin.  The total number of negative examples is
controlled by the size of the sets $\cN_E$ and $\cN_T$. We experiment by sampling only entities or only types or both by fixing the total
number of negative examples in Section~\ref{sec:experiments}.

The rest of section is organized as follows: we propose three
algorithms based on the global objective in
Section~\ref{sec:algorithms}.  In Section~\ref{sec:relat-exist-meth},
we discuss the relationship between the proposed algorithms and
existing approaches.  Let $\Phi(e)
\rightarrow R^{d_{e}}$ be the feature function that maps an entity to
its feature representation, and $\Psi(t) \rightarrow R^{d_{t}}$ be
the feature function that maps an entity type to its feature
representation.\footnote{This gives the possibility of defining
  features for the labels in the output space but we use a simple
  one-hot representation for types right now since richer features did
  not give performance gains in our initial experiments.}
$d_e$ and $d_t$ represent the feature dimensionality of the entity
features and the type features respectively. Feature representations of the entity  types ($\Psi$) is
only used in the embedding model.

\subsection{Algorithms}
\label{sec:algorithms}
We propose three different algorithms based on the
global objective framework for predicting missing entity types. Two
algorithms use the linear  model and the other one uses the
embedding model.

\paragraph{Linear Model}
The scoring function in this model is given by $s(e,
t|\theta\!=\!\{\bw_{t}\}) = \bw_{t}^T\Phi(e)$, where
$\bw_{t} \in R^{d_{e}} $ is the parameter vector for target type
$t$. The regularization term in Eq.~\eqref{eq:general_obj} is defined
as follows: $R(\theta) = 1/2 \sum_{t=1} \bw_t^T\bw_t$. We use $k=2$ in
our experiments. Our first algorithm is obtained by using the dual
coordinate descent algorithm~\cite{dcd} to optimize
Eq.~\eqref{eq:general_obj}, where we modified the original algorithm
to handle multiple weight vectors. We refer to this algorithm as {\bf
  Linear.DCD}.

While DCD algorithm ensures convergence to the global
optimum solution, its convergence can be slow in certain cases. Therefore,
 we adopt an online algorithm, Adagrad~\cite{adagrad}. We use the hinge loss
function ($k=1$) with no
regularization ($Reg(\theta) = \emptyset$) since it gave best results in our initial experiments. We refer to this algorithm as {\bf Linear.Adagrad},
which is described in Algorithm~\ref{algo_adagrad}. Note that $\text{AdaGradUpdate}(x,g)$ is a procedure which updates the vector $x$ with the respect to the gradient $g$.
%Note that
%we call the function AdaGradUpdate($\bw$,$\bg$) in line 5,10 and
%11. The function updates the weight vectors with the gradient
%nformation and makes use of historical gradient information to adjust
%the learning rate. More precisely, AdaGradUpdate iterates through the
%feature index $i$ and performs the following steps:
%\begin{equation*}
 % \begin{split}
   % \bh[i] &\leftarrow \bh[i] + \bg[i]^2, \\
    %\bw[i] &\leftarrow \bw[i] - \eta \bg[i] / \sqrt{\bh[i]},
 % \end{split}
%\end{equation*}
%where $\bh$ stores the historical gradient information. Note that we use
%$\bw[i]$ to represent the $i$-th element of the vector $\bw$.

\newfloat{algorithm}{t}{lop}
\begin{algorithm}
\footnotesize
\caption{ The training algorithm for Linear.Adagrad. }
\label{algo_adagrad}
\begin{algorithmic}[1]
 \State Initialize $\bw_t=0, \forall t=1\ldots |T|$
 \For{$(e, t) \in \Lambda_0$ }
      \For{$e' \in \cN_E(e,t)$ }
      \If{$\bw_{t}^T\Phi(e) - \bw_{t}^T\Phi(e') -1 < 0$ }
      \State $\text{AdaGradUpdate}(w_t, \Phi(e') - \Phi(e))$
      \EndIf
      \EndFor
      \For{$t' \in \cN_T(e,t)$ }
      \If{$\bw_{t}^T\Phi(e) - \bw_{t'}^T\Phi(e) -1 < 0$ }
      \State $\text{AdaGradUpdate}(w_t, -\Phi(e))$
      \State $\text{AdaGradUpdate}(w_{t'},\Phi(e))$.
      \EndIf
      \EndFor
 \EndFor
%\State Output: $\{\bw_t\}$
\end{algorithmic}
\end{algorithm}

\paragraph{Embedding Model}
\newfloat{algorithm}{t}{lop}
\begin{algorithm}
\footnotesize
\caption{The training algorithm for the embedding model.}
\label{algo:embedding}
\begin{algorithmic}[1]
% \State Input: $\Lambda_0$, $d$, $\Phi(e)$, $\Psi(t)$
 \State Initialize $\bV,\bU$ randomly.
 \For{$(e, t) \in \Lambda_0$ }
     \For{$e' \in \cN_E(e,t)$ }
         \If{$s(e,t) - s(e',t) -1 < 0$ }
             \State $\mu \leftarrow \bV^T \Psi(t)$
             \State $\eta \leftarrow \bU^T (\Phi(e') - \Phi(e))$
             \For{ $i \in 1\ldots d$}
               \State $\text{AdaGradUpdate}(\bU_i, \mu[i] (\Phi(e')- \Phi(e)))$
               \State $\text{AdaGradUpdate}(\bV_i, \eta[i] \Psi(t))$
             \EndFor
         \EndIf
      \EndFor
      \For{$t' \in \cN_T(e,t)$ }
          \If{$s(e,t) - s(e,t') -1 < 0$ }
              \State $\mu \leftarrow \bV^T (\Psi(t') - \Psi(t))$
              \State $\eta \leftarrow \bU^T\Phi(e)$
              \For{ $i \in 1\ldots d$}
                  \State $\text{AdaGradUpdate}(\bU_i, \mu[i] \Phi(e))$
                  \State $\text{AdaGradUpdate}(\bV_i, \eta[i] (\Psi(t') - \Psi(t)))$
              \EndFor
          \EndIf
      \EndFor
 \EndFor
%\State Output: $V^T, \bU^T$
%
\end{algorithmic}
\end{algorithm}

In this model, vector representations are constructed for entities and
types using linear projection matrices. Recall $\Psi(t)
\rightarrow R^{d_{t}}$ is the feature function that maps a type to its
feature representation. The scoring function
is given by
\begin{center} $ s(e, t|\theta\! =(\bU,\bV)) = \Psi(t)^T\bV\bU^T
  \Phi(e), $ \end{center} where $\bU \in R^{d_e \times d}$ and $\bV
\in R^{d_{t}\times d}$ are projection matrices that embed the entities
and types in a $d$-dimensional space.  Similarly to the linear classifier model,  we use the l1-hinge loss
function ($k=1$) with no
regularization ($Reg(\theta) = \emptyset$).  $\bU_i$ and $\bV_i$ denote the $i$-th column vector of the matrix $\bU$ and $\bV$,
respectively.  The algorithm is described in detail in
Algorithm~\ref{algo:embedding}.

The embedding model has more expressive power than the linear
 model, but the training unlike in the linear
model, converges only to a local optimum solution since the objective
function is non-convex.

\subsection{Relationship to Existing Methods}
\label{sec:relat-exist-meth}

% Prior to this work, there are {\em type-based approaches}, which
% sample negative examples only from the type side
% ~\cite{distant_supervision,limin} and {\em entity-based approaches},
% that sample negative examples only from the entity
% side~\cite{fine-grained}. The global training objective approach
% combines the advantages of these two approaches by using negative
% examples from both the entity and the type side.
Many existing methods for relation extraction and entity type
prediction can be cast as a special case under the global objective
framework. For example, we can consider the
work in relation extraction~\cite{distant_supervision,transe,limin} as  models trained with
$\cN_T(e,t) = \emptyset$. These models are trained only using negative entities which we refer to as Negative Entity (NE) objective. The entity type prediction model in \newcite{fine-grained} is a
linear model with $\cN_E(e,t) = \emptyset$ which we refer to as the Negative Type (NT) objective.
The embedding model described in \newcite{wsabie} developed for image
retrieval is also a special case of our model trained with the NT objective.

While the $NE$ or $NT$ objective functions could be suitable for some
classification tasks \cite{wsabie}, the choice of objective functions
for the KBC tasks has not been well motivated. Often the choice is made neither with
theoretical foundation nor with empirical support. To the best of our
knowledge, the global objective function, which includes both
$\cN_E(e,t)$ and $\cN_T(e,t)$, has not been considered previously by
KBC methods. % The global objective

\section{Experiments}
\label{sec:experiments}

\begin{table}[t!]
\centering
\small
  \begin{tabular}{ l | c |c }
    \hline
    & 70 types & 500 types \\  \hline \hline
    Entities & 2.2M  & 2.2M \\\hline
    \multicolumn{3}{c}{Training Data Statistics ($\Lambda_0$)}\\ \hline \hline
    positive example & 4.5M & 6.2M \\
    max \#ent for a type  & 1.1M & 1.1M \\
    min \#ent for a type  & 6732 & 32 \\\hline
    \multicolumn{3}{c}{Test Data Statistics ($\Lambda - \Lambda_0$)}\\ \hline \hline
    positive examples & 163K & 240K \\
    negative examples & 17.1M & 132M \\
    negative/positive ratio &  105.22 & 554.44 \\\hline
  \end{tabular}
  \caption{\small Statistics of our dataset. $\Lambda_0$ is our training snapshot and $\Lambda$ is
    our test snapshot. An example is an entity-type pair.}
  \label{table:stats}
\end{table}
In this section, we give details about our dataset and discuss our experimental results. Finally, we perform manual evaluation on a small subset of the data.

\subsection{Data}
First, we evaluate our methods on $70$ entity types with the most
observed facts in the training data.\footnote{We removed few entity
  types that were trivial to predict in the test data.}  We also
perform large-scale evaluation by testing the methods on $500$ types
with the most observed facts in the training data.

Table \ref{table:stats} shows statistics of our dataset. The number of
positive examples is much larger in the training data compared to
that in the test data since the test set contains only facts that were
added to the more recent snapshot. An additional effect of this is that most of the facts in the test data are about entities that are not  very {\emph well-known or famous}.
The high negative to positive examples ratio in the test data makes this dataset very challenging.

\begin{table*}[t!]
\small
\begin{subtable}{.47\linewidth}
\centering
\begin{tabular}{ l | l | l | l  }
\hline
Features & Algorithm & MAP & GAP  \\  \hline
 \multirow{2}{*} {Description} & Linear.Adagrad  & 29.17 & 28.17 \\
 & Linear.DCD  & 28.40 & 27.76 \\\hline
 \multirow{2}{*} {\begin{minipage}{1in} Description + Wikipedia \end{minipage}} & Linear.Adagrad  & {\bf 33.28} & {\bf 31.97} \\
 & Linear.DCD  & 31.92 & 31.36 \\\hline
\end{tabular}
\caption{Adagrad vs. Dual coordinate descent (DCD). Results are obtained using linear models trained with global training objective (m=1, n=1) on 70 types.  }\label{table:train}
\end{subtable}
\begin{subtable}{.47\linewidth}
\centering
\begin{tabular}{ l | l | l  }
\hline
Features & MAP & GAP  \\  \hline
Type (T) & 12.33  & 13.58 \\\hline
Description (D) & 29.17 & 28.17 \\\hline
Wikipedia (W) & 30.81  & 30.56 \\\hline
D + W & 33.28 & {\bf 31.97} \\\hline
T + D + W & {\bf 36.13}  & 31.13 \\\hline
\end{tabular}
\caption{Feature Comparison. Results are obtained from using Linear.Adagrad with global training objective (m=1, n=1) on 70 types.}\label{table:feature}
\end{subtable}
\begin{subtable}{\linewidth}
\centering
\begin{tabular}{ l  l  l  l  }
& & &  \\
& & &  \\
\end{tabular}
\end{subtable}
\begin{subtable}{0.47\linewidth}
\centering
\begin{tabular}{ l | l | l | l   }
\hline
Features  & Objective & MAP & GAP \\  \hline
\multirow{3}{*} { D + W } & NE (m = 2)  & 33.01 & 23.97   \\
& NT (n = 2)  & 31.61 & 29.09   \\
& Global (m = 1, n = 1)  & 33.28 & {\bf 31.97}   \\\hline
\multirow{3}{*} {T + D + W} & NE (m = 2)  & 34.56       & 21.79    \\
& NT (n = 2)  & 34.45       & 31.42    \\
& Global (m = 1, n = 1)  & {\bf 36.13}  & 31.13  \\\hline
\end{tabular}
\caption{Global Objective vs NE and NT. Results are obtained using Linear.Adagrad on 70 types.}\label{table:objective}
\end{subtable}
\begin{subtable}{0.47\linewidth}
\centering
\begin{tabular}{ l | l | l | l   }
\hline
Features  & Objective & MAP & GAP \\  \hline
\multirow{3}{*} { D + W } & NE (m = 2)   & 30.92          & 22.38   \\
& NT (n = 2)  & 25.77          & 23.40   \\
& Global (m = 1, n = 1)  & {\bf 31.60}          & {\bf 30.13}    \\\hline
\multirow{3}{*} {T + D + W} & NE (m = 2) & 28.70          & 19.34   \\
& NT (n = 2)  & 28.06          & 25.42    \\
& Global (m = 1, n = 1)  & 30.35          & 28.71  \\\hline
\end{tabular}
\caption{Global Objective vs NE and NT. Results are obtained using the embedding model on 70 types.}
\label{table:embeddingobjective}
\end{subtable}
\begin{subtable}{\linewidth}
\centering
\begin{tabular}{ l  l  l  l  }
& & &  \\
& & &  \\
\end{tabular}
\end{subtable}
\begin{subtable}{\linewidth}
\centering
\begin{tabular}{ l | l | l | l | l | l  }
\hline
Features  & Model & MAP & GAP & G@1000 & G@10000  \\  \hline
\multirow{2}{*} {D + W} & Linear.Adagrad & 33.28 &  {\bf 31.97}  & {\bf 79.63} & {\bf 68.08}\\
& Embedding  & 31.60 & 30.13 & 73.40 & 64.69   \\\hline
\multirow{2}{*} {T + D + W} & Linear.Adagrad  & {\bf 36.13}  & 31.13 & 70.02 & 65.09 \\
& Embedding  & 30.35 & 28.71 & 62.61 & 64.30   \\\hline
\end{tabular}
\caption{Model Comparison. The models were trained with the global training objective (m=1, n=1) on 70 types.}\label{table:model}
\end{subtable}
\begin{subtable}{\linewidth}
\centering
\begin{tabular}{ l  l  l  l  }
& & &  \\
& & &  \\
\end{tabular}
\end{subtable}
\begin{subtable}{\linewidth}
\centering
\begin{tabular}{   l | l | l | l |l }
\hline
Model & MAP & GAP    & G@1000 & G@10000  \\  \hline
Linear.Adagrad & {\bf 13.28} & {\bf 20.49} & {\bf 69.23}  & {\bf 60.14}  \\ \hline
Embedding & 9.82 & 17.67 & 55.31  & 51.29  \\ \hline
\end{tabular}
\caption{Results on 500 types using Freebase description features. We train the models with the global training objective (m=1, n=1).}\label{table:full}
\end{subtable}
\caption{ \small Automatic Evaluation Results. Note that $m = |\cN_E (e,t)|$ and $n = |\cN_T (e,t)|$.}\label{table:small}
\end{table*}

\subsection{Automatic Evaluation Results}
Table \ref{table:small} shows automatic evaluation results where we give results on $70$ types  and $500$ types. We compare  different aspects of the system on $70$ types empirically.
\paragraph{Adagrad Vs DCD}
We first study the linear models by comparing Linear.DCD and Linear.AdaGrad.  Table \ref{table:train} shows that Linear.AdaGrad consistently performs better for our task.

 \paragraph{Impact of Features}

We compare the effect of different features on the final performance using
  Linear.AdaGrad in Table \ref{table:feature}. Types are represented by boolean features while Freebase description and Wikipedia full text are represented using \emph{tf-idf} weighting. The best MAP results are obtained by using all the information (T+D+W) while best GAP results are obtained by using the Freebase description and Wikipedia article of the entity. Note that the features are simply concatenated when multiple resources are used. We tried to use \emph{idf} weighting on type features and on all features, but they did not yield improvements.

\paragraph{The Importance of Global Objective}
Table \ref{table:objective} and  \ref{table:embeddingobjective} compares global training objective with NE and NT training objective. Note that all the three methods use the same number of negative examples. More precisely, for
each $(e,t) \in \Lambda_0$, $|\cN_E(e,t)| + |\cN_T(e,t)| = m+ n =  2$. The results show that the global training objective achieves best scores on both MAP and GAP for classifiers and  low-dimensional embedding models. Among NE and NT, NE performs better on the type-based metric while NT performs better on the global metric.

\paragraph{Linear Model Vs Embedding Model}
Finally, we compare the linear classifier model with the embedding model in Table \ref{table:model}. The linear classifier model performs better than the embedding model  in both MAP and GAP.

We perform large-scale evaluation on $500$ types  with the description features (as  experiments are
 expensive) and the results are shown in Table \ref{table:full}.  One might expect that with the increased number of types, the embedding
model would perform better than the classifier since they share parameters across types. However, despite the recent popularity of embedding models in NLP, linear model still performs better in our task.

\subsection{Human Evaluation}
To verify the effectiveness of our KBC algorithms, and the correctness of our automatic evaluation method, we perform
manual evaluation on the top $100$ predictions of the output obtained
from two different experimental setting and the results are shown in
Table \ref{table:manual}. Even though the automatic evaluation gives
pessimistic results since the test KB is also incomplete\footnote{This
  is true even with existing automatic evaluation methods.}, the
results indicate that the automatic evaluation is correlated with
manual evaluation. More excitingly, among the 179 unique instances we
manually evaluated, 17 of them are still\footnote{at submission time.} missing in Freebase which
emphasizes the effectiveness of our approach.

\subsection{Error Analysis}
\begin{itemize}
\item{ {\bf Effect of training data}: We find the performance of the models on a type is highly dependent on the number of training instances for that type. For example, the linear classifier model  when evaluated on $70$ types performs 24.86 \% better on the most frequent $35$ types compared to the least frequent $35$ types.
This indicates  bootstrapping or active learning techniques can be profitably used to provide more supervision for the methods. In this case, G@k would be an useful metric to compare the effectiveness of the different methods.   }
\item{ {\bf Shallow Linguistic features}:  We found some of the false positive predictions are caused by the use of shallow linguistic features. For example, an  entity who has acted in a movie and composes music only for television shows is wrongly tagged with the type {\emph /film/composer} since words like "movie", "composer" and "music" occur frequently in the Wikipedia article of the entity (\url{http://en.wikipedia.org/wiki/J._J._Abrams}).  }
\end{itemize}
\begin{table}[t!]
\small
  \begin{tabular}{ l | l | l | l  }
    \hline
    Features & G@100 & G@100-M & Accuracy-M \\\hline
    D + W & {\bf 87.68} & {\bf 97.31} & {\bf 97}  \\ \hline
    T + D + W & 84.91 & 91.47 & 88 \\ \hline
  \end{tabular}
  \caption{\small Manual vs. Automatic evaluation of top 100 predictions on 70 types. Predictions are obtained by training a linear classifier using Adagrad with global training objective (m=1, n=1). G@100-M and Accuracy-M are computed by manual evaluation.}
  \label{table:manual}
\end{table}

%%% Local Variables:
%%% mode: latex
%%% TeX-master: "paper"
%%% TeX-PDF-mode: t
%%% End:

\section{Related Work}
\label{sec:related-work}

\paragraph{Entity Type Prediction and Wikipedia Features} Much of previous work \cite{pantel,fine-grained} in entity type prediction has focused on the task of predicting entity types at the sentence level. \newcite{uschema} develop a method based on matrix factorization for entity type prediction in a KB using information within the KB and New York Times articles. However, the method was still evaluated only at the sentence level. \newcite{toral}, \newcite{wiki} use the first line of an entity's  Wikipedia article to perform named entity recognition on three entity types.

\paragraph{Knowledge Base Completion} Much of precious work in KB completion has focused on the problem of relation extraction. Majority of the methods infer missing relation facts  using information within the KB \cite{rescal,pra,rntn,transe} while methods such as \newcite{distant_supervision} use information in text documents. \newcite{limin} use both information within and outside the KB to complete the KB.

\paragraph{Linear Embedding Model} \newcite{wsabie} is one of first work that developed a  supervised linear embedding model and applied it to image retrieval. We apply this model to entity type prediction but we train using a different objective function which is more suited for our task.

%%% Local Variables:
%%% mode: latex
%%% TeX-master: "paper"
%%% TeX-PDF-mode: t
%%% End: 

\section{Conclusion and Future Work}
\label{sec:concl-future-work}

We propose an evaluation framework comprising of  methods for dataset construction and evaluation metrics to evaluate KBC approaches for inferring  missing entity type instances. We verified that our automatic evaluation is correlated with human evaluation, and our dataset and evaluation scripts are publicly available.\footnote{ \url{http://research.microsoft.com/en-US/downloads/df481862-65cc-4b05-886c-acc181ad07bb/default.aspx}} Experimental results show that models trained with our proposed global training objective produces higher quality ranking within and across types when compared to baseline methods.

In future work, we plan to use information from entity linked documents to improve performance and also explore active leaning, and other human-in-the-loop methods to get more training data.

%%% Local Variables:
%%% mode: latex
%%% TeX-master: "paper"
%%% TeX-PDF-mode: t
%%% End:

\bibliographystyle{naaclhlt2015}
\bibliography{naacl}

\end{document}